\documentclass{article}

\usepackage{graphicx}
\usepackage{arxiv}
\usepackage[utf8]{inputenc} 
\usepackage[T1]{fontenc}    
\usepackage{scalerel}
\usepackage{tikz}

\usepackage{hyperref}       
\usepackage{url}            
\usepackage{booktabs}       
\usepackage{amsfonts}       
\usepackage{nicefrac}       
\usepackage{microtype}      
\usepackage{lipsum}

\usepackage{amsmath, mathrsfs, amssymb}
\usepackage{cleveref}
\usepackage[sort&compress,numbers]{natbib}
\usepackage{subfigure}
\usepackage{algorithm, algorithmic}
\usepackage{comment}

\usepackage{xcolor}
\usepackage{tikz}
\usetikzlibrary{svg.path}

\title{Out-of-Bag Anomaly Detection}
\author{
\href{https://orcid.org/0000-0002-0874-0578}{\includegraphics[scale=0.06]{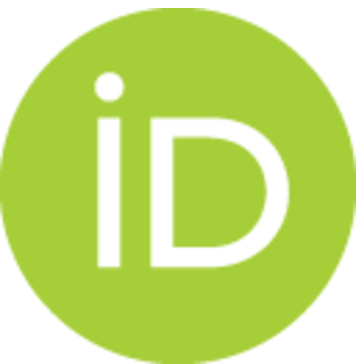}\hspace{1mm}}
Egor Klevak \\
Zillow Group \\
Seattle, WA \\
\texttt{egork@zillowgroup.com}\\
\And
\href{https://orcid.org/0000-0003-2277-3701}{\includegraphics[scale=0.06]{orcid.eps}\hspace{1mm}}
Sangdi Lin\\
Zillow Group \\
Seattle, WA \\
\texttt{sangdil@zillowgroup.com}\\
\And
Andy Martin\\
Zillow Group \\
Seattle, WA \\
\texttt{andyma@zillowgroup.com}\\
\And
Ondrej Linda\\
Zillow Group \\
Seattle, WA \\
\texttt{ondrejl@zillowgroup.com}\\
\And
Eric Ringger\\
Zillow Group \\
Seattle, WA \\
\texttt{ericri@zillowgroup.com}\\
}
\begin{document}
\maketitle

\begin{abstract}
Data anomalies are ubiquitous in real world datasets, and can have an adverse impact on machine learning (ML) systems, such as automated home valuation. Detecting anomalies could make ML applications more responsible and trustworthy. However, the lack of labels for anomalies and the complex nature of real-world datasets make anomaly detection a challenging unsupervised learning problem. In this paper, we propose a novel model-based anomaly detection method, that we call Out-of-Bag anomaly detection, which handles multi-dimensional datasets consisting of numerical and categorical features. The proposed method decomposes the unsupervised problem into the training of a set of ensemble models. Out-of-Bag estimates are leveraged to derive an effective measure for anomaly detection. We not only demonstrate the state-of-the-art performance of our method through comprehensive experiments on benchmark datasets, but also show our model can improve the accuracy and reliability of an ML system as data pre-processing step via a case study on home valuation.
\end{abstract}


\section{Introduction}
\label{sec:intro}


Data anomalies are very common in real world datasets. Some anomalies are generated by errors made in data recording, which occurs often in transactional data in E-commerce. Others could imply fraudulent information on the web which has an adverse impact on people's trust on  web applications. Anomalies can sometimes be the indicators of potential risks such as fraudulent transactions in credit card processing \citep{hodge2004survey, domingues2018comparative}, and identifying those would make the digital world a safer and more trustworthy place.

\begin{figure}[t]
\begin{center}
    \includegraphics[width=0.8\textwidth]{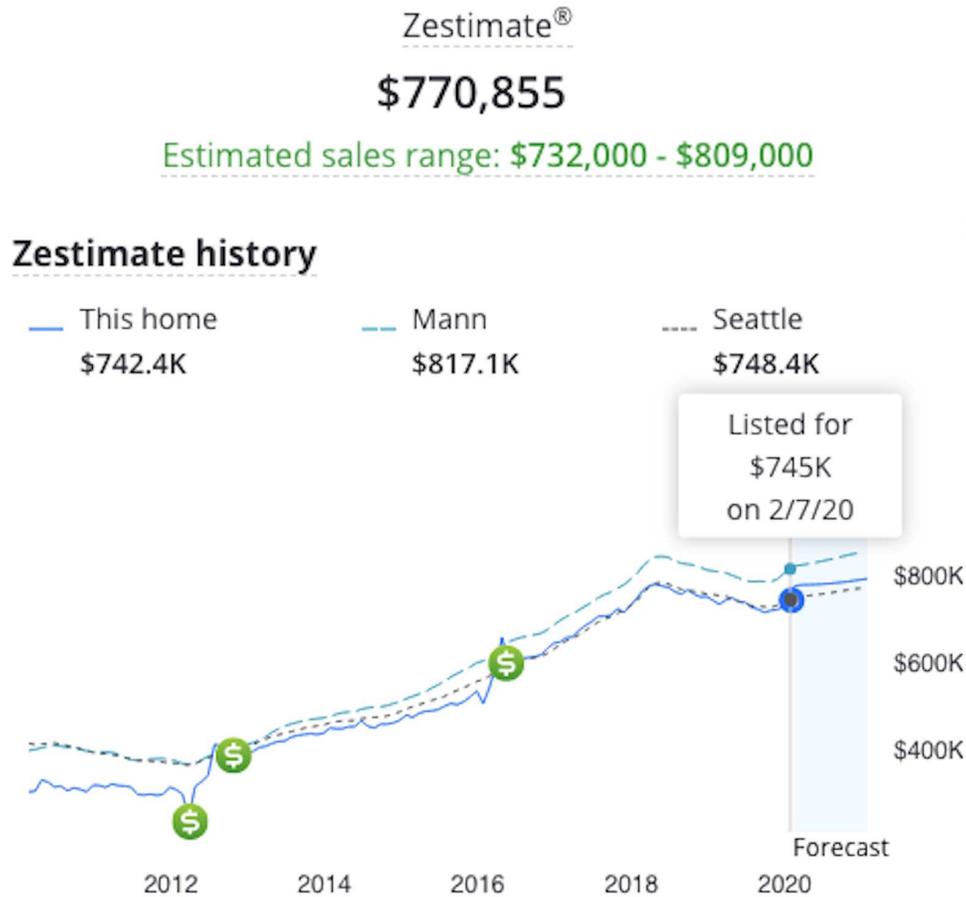}
    \caption{Example of Zestimate\textregistered}
    \label{fig:zestimate_example}
\end{center}
\end{figure}

In machine learning (ML), anomalies are observations that are distinct from the majority of the data population and are typically generated by sufficiently different generating processes. Machine learning models are often used to detect anomalies by profiling the normal data distribution or data generating processes. To build a trusted ML application, anomaly or outlier detection becomes a key component of the data preparation pipeline. For example, in home-buying, automated home-valuation systems such as the Zestimate\textregistered \cite{zestimate} as shown in \cref{fig:zestimate_example} empower people with information that has fundamentally transformed the real-estate industry. In order to support 
the biggest transaction in a customer's life with trusted information, it is important to have an unbiased and reliable ML system for predicting home value. The reliability of the transaction data influences the accuracy of such ML system. 

One approach to identify anomalies is through supervised learning, which would require tremendous efforts of human labeling for any industrial scale databases. It is also not feasible when applications require diverse sources of data that often come in disparate formats. Therefore, Unsupervised techniques are the preferred solution in such scenarios. Moreover, real-world datasets often have high dimensionality and mixed feature types (numerical and categorical), which imposes a challenging task for anomaly detection. For example, US real estate transactional data has non-homogeneous sources for different markets with variable definitions for anomalies. Most of the sources have humans in the loop, and errors such as mistyping the number of zeros or transposing the order of the digits are quite common. Such mistakes not only influence the performance of the ML models, but also cause long lasting real-life impact \citep{desertnews2019}. Anomaly detection techniques play an important role in supporting our decision making with trusted information in the era of big data.

In this paper, we propose a novel anomaly detection method, that we call Out-of-Bag (OOB) anomaly detection. The proposed method first converts the unsupervised problem into a set of supervised problems, where an ensemble model is trained to model each feature. Next, the method utilizes the OOB predictions of ensemble models to derive important statistics for detecting anomalous data points. More specifically, we developed two important measures for scoring anomalies: an uncertainty-based measure and a disagreement-based measure. Both measures can be combined into an effective anomaly score to identify various types of anomalous data. The OOB anomaly detection method demonstrates the state-of-art performance as demonstrated via an extensive set of experiments on benchmark datasets. In addition, we show that our method significantly improves the accuracy and reliability of the ML model by removing the detected anomalies from our in-house home valuation dataset. 

The remainder of this paper is structured as follows. After discussing related work in \cref{sec:related}, we present the details of the OOB anomaly detection method in \cref{sec:method}. In \cref{sec:results}, we first show that the proposed method demonstrates the state-of-the-art performance of anomaly detection compared to other competing methods through an extensive set of experiments on benchmark datasets. In addition, it is shown a more effective pre-processing approach to filter out anomalous training data in an in-house ML home valuation application. Finally, we conclude the paper in \cref{sec:conclusion}.

\section{Related Work}
\label{sec:related}

Anomaly detection previouslt has been addressed by several different approaches. The existing techniques roughly fall under 4 categories: (1) distance-based methods, (2) reconstruction based methods, (3) probability-based methods and (4) model-based methods. 

Distance-based methods use the proximity relationship to other data points to identify anomalous data points. For example, \citep{ramaswamy2000efficient} uses the distance to the k-nearest neighbor as the anomaly score. However, distance-based methods often suffer from the “curse of dimensionality”, as distance between data points in high dimensional space becomes similar. Angle-Based Outlier Detection (ABOD) \citep{kriegel2008angle} is proposed to mitigate the effect of high dimensionality by considering a outlier factor measured by variance of the angles to other data points. Clustering techniques have also been used to detect anomalies as a side effect of cluster discovery in the data. Cluster-Based Local Outlier Factor (CBLOF) \citep{he2003discovering} considers the distance to other data as well as the the size of the clusters in computing anomaly scores. 

Reconstruction-based methods learn the principle factors in the data and a reconstruction function to reconstruct the data from these principle factors. In this process, anomalous data points can be identified by high reconstruction error. Principal Component Analysis (PCA)-based methods \citep{shyu2006principal} are able to learn the linear relationship between features, while Autoencoder (AE)-based methods \citep{kramer1991nonlinear, kingma2013auto} can handle the non-linear interactions in the data more efficiently. 

Probability-based approaches fit the normal data by a probability distribution. The distribution can be estimated by parametric models such as Gaussian Mixture Models (GMMs) \citep{dempster1977maximum} or non-parametric models such as the feature histograms used in Histogram-based Outlier Score (HBOS) \citep{goldstein2012histogram}. Anomalies are identified based on the likelihood of occurrence in the estimated distribution which profiles the normal data. Moreover, some Variational Autoencoder  (VAE)-based methods \citep{an2015variational, zimmerer2018context} combine probability- with reconstruction-based techniques in scoring anomalies.

Model-based methods cover a broad spectrum of methods. In general, models are taught to profile the normal pattern and feature relationship in the data. They often involve a conversion from the original unsupervised setting into a supervised setting so that supervised techniques can be applied. For example, One-Class Support Vector Machine (OC-SVM) \citep{scholkopf2000support} finds a decision boundary that maximizes the distance between data points to the origin in the feature space to differentiate anomalies from the normal data. Isolation Forest (IF) \citep{isolation_forest}, another model-based method, utilizes an ensemble of random trees which partition on random features and values to isolate the anomalies. Anomalies can be identified by a shorter average path length as their infrequent feature values require fewer splits to isolate. As a tree-based method, IF also benefits from the advantage of handling  high-dimensional data.

An important concept which we use in the proposed approach is Out-of-Bag Models. First introduced by Breiaman \citep{breimanOOB}, bootstrap aggregation of training data in ensemble methods \citep{breimanBagging} results in approximately 33 percent of learners not using specific data point for training. The Out-of-Bag (OOB) models have been used to estimate the generalization error of Random Forest ensemble models \citep{tibshirani1996bias, wolpert1999efficient}. Empirical evidence demonstrates that the OOB estimate of generalized error is accurate and unbiased \citep{breiman2001random}. In our method, we utilize OOB estimates to derive the measure for anomalies.
\section{Method}
\label{sec:method}

In this paper, we propose a novel anomaly detection method which solves the challenging unsupervised problem from a supervised perspective. At a high level, we train a set of ensemble models, and each model is targeted to predict one feature column given the remaining features. The concept of OOB predictions in ensemble models is employed in developing the score for measuring the degree of anomaly. We discuss how we can handle both categorical and numerical features under the proposed framework, and extensively elaborate the intuition behind the proposed method. The notations for the proposed method are summarized in \cref{tab:notations}, and \cref{alg:algorithm} summarizes the proposed method. 
\subsection{Methodology Overview}
\label{sec:method_details}

\begin{table}[h!]
  \begin{center}
    \caption{Summary of notations}
    \label{tab:notations}
    \begin{tabular}{c|c}
      \textbf{Symbol} & \textbf{ Meaning} \\
     \hline
     $N$ & number of data points \\
     $K$ & number of features \\
     $T$ & number of models in each ensemble \\
      $x_i^k$& the $k$-th feature value of the $i$-th data point\\
      $\bold{x}_i = (x_i^1, x_i^2, \ldots, x_i^K)$  & the $i$-th data point (row) \\
      $\bold{x}^k = (x_1^k, x_2^k, \ldots, x_N^k)$ & the $k$-th feature (column) \\
      $\bold{x}^{-k}$  &  feature columns excluding $\bold{x}^k$ \\
      $S_i$ & anomaly score for $\bold{x}_i$\\ 
      $S_i^k$ & anomaly score for $x_i^k$ \\
      $S_i^k (\textrm{uncertainty})$ & uncertainty measure for $x_i^k$ \\
      $S_i^k(\textrm{disagreement})$ & disagreement measure for $x_i^k$ \\
      $\bold{M}^k$ & ensemble model: $\bold{x}^{-k} \rightarrow \bold{x}^k$ \\
      $T_i^k$ & number of OOB models for $\bold{x}_i$  in $\bold{M}^k$ \\
       $ \hat{\bold{x}}_i^k $ & OOB predictions for $x_i^k$ \\
       $\hat{x}_i^k(t)$ & the $t$-th OOB prediction for $x_i^k$ \\
       $\bar{x}_{i}^k $ & average of OOB predictions   $ \hat{\bold{x}}_i^k $  (numerical)\\ 
      $C_k$ & cardinality of $\bold{x}^k$ if categorical
    \end{tabular}
  \end{center}
\end{table}

Let $\{x_i^k \}_{i = 1, \ldots, N; k= 1, \ldots, K}$ represent a dataset with $N$ data points and $K$ features.
We use $\bold{x}_i = (x_i^1, x_i^2, \ldots, x_i^K)$ to represent the $i$-th data point (i.e., row or feature vector) each including $K$ features, and $\bold{x}^k = (x_1^k, x_2^k, \ldots, x_N^k)$ to represent the $k$-th feature column of the dataset. In addition, we define $\bold{x}^{-k}$ $=$ $(\bold{x}^1$, \ldots, $\bold{x}^{k-1}, \bold{x}^{k+1}, \ldots, \bold{x}^K)$ to represent the set of feature columns excluding the $k$-th feature. Similarly, for the $i$-th data point, $\bold{x}_i^{-k} = (x_i^1, \ldots, x_i^{k-1}, x_i^{k+1}, \ldots, x_i^K)$ represents the $i$-th feature vector excluding its $k$-th feature.

For each data point $\bold{x}_i$, we calculate an anomaly score $S_i$ as the sum over the anomaly score of each of its features denoted by $S_i^k, k = 1, \ldots, K$ (\cref{eq:agg_func_avg}). The higher the score is, the more likely that the data point is an anomaly.

\begin{equation}\label{eq:agg_func_avg}
S_i = \sum_{k=1}^{K} S_i^k, i = 1, \ldots, N.
\end{equation}

To calculate $S_i^k$, we train an ensemble model $\bold{M}^k$ with $T$ base learners $\{ M_1^k, M_2^k, \ldots,M_T^k\}$ to predict feature column $\bold{x}^k$ 
using the remaining feature columns $\bold{x}^{-k}$.  Each base learner is trained on a bootstrap sample of the full dataset. Formally speaking, we have
\begin{equation}
    \bold{M}^k: \bold{x}^{-k} \rightarrow \bold{x}^k, k = 1, \ldots, K.
\end{equation}

We use the Random Forest \cite{breiman2001random} as our ensemble model where a base learner is either a decision tree for a categorical feature or regression tree for a numerical feature. Note that the proposed method is generally applicable to all kinds of classification or regression models. We choose tree-based models for their intrinsic ability to handle high dimensionality and features with disparate types and scales.

In an ensemble model,  OOB predictions for a data point are defined as the predictions made by the set of base models which do not include that data point as part of the training set due to the bootstrap sampling.  We also call these models OOB models for that data point.
Let $\bold{\hat{x}} _i^k =  \{ \hat{x} _i^k(1),  \ldots,  \hat{x} _i^k(T_i^k) \} $ represent the OOB predictions from the OOB models with respect to $x_i^k$ , where $T_i^k(<T)$ is the number of such OOB models for the $i$-th data point in the ensemble model $\bold{M}^k$.

The anomaly score $S_i^k$ is calculated as a sum of two derived statistics from the OOB predictions (\cref{eq:two_terms}). The first term describes the uncertainty in the OOB predictions, and the second term describes the disagreement between $x_i^k$ and the expected value by consulting the OOB models. 
\begin{equation}
\label{eq:two_terms}
    S_i^k = S_i^k(\mathrm{uncertainty}) + S_i^k (\mathrm{disagreement}), k = 1, \ldots, K, i = 1, \ldots, N.
\end{equation}

Below, we discuss in detail how  $S_i^k(\mathrm{uncertainty}) $ and $S_i^k (\mathrm{disagreement}) $ are computed from the OOB predictions in the cases of a categorical feature and a numerical feature, respectively.

\subsection{Categorical Features}
\label{sec:cat_feature}
When feature $\bold{x}^k $ is a categorical feature of cardinality $C_k$,  each element of the OOB prediction vector $\bold{\hat{x}} _i^k = \{ \hat{x} _i^k(1),  \ldots,  \hat{x} _i^k(T_i^k))  \} $ points to one of the possible categorical values. A natural choice of the uncertainty measure is entropy:

\begin{equation}
\label{eq:cat_uncertainty}
S_i^k(\textrm{uncertainty}) =\frac{1}{\log C_k} Entropy( \hat{x} _i^k(1),  \ldots,  \hat{x} _i^k(T_i^k)) ).
\end{equation}
The scaling factor ($=1/\log C_k$) in \cref{eq:cat_uncertainty}  is the maximum entropy for a $C_k$-cardinality distribution.

To characterize the disagreement, we simply calculated the difference between 1 and the expected probability of $x_i^k$ judged by the OOB models:
\begin{equation}
S_i^k(\textrm{disagreement}) = 1 - \frac{\sum_{t=1}^{T_i^k} \bold{I}(x_i^k(t) == x_i^k)}{T_i^k},
\end{equation}
where $\bold{I}$ (condition) is a function that outputs 1 if the condition is satisfied and 0 otherwise.

\subsection{Numerical Features}
\label{sec:num_feature}
In cases where $\bold{x}^k$ is a numerical feature, anomaly score can be simply computed as the average squared difference between the OOB predictions and the feature value:

\begin{equation}
\label{eq:num_anomaly}
S_i^k = \frac{1}{T_i^k} \sum_{t=1}^{T_i^k} (\hat{x}_i^k(t) - x_i^k  )^2.
\end{equation}
\Cref{eq:num_anomaly} provides a combined view of  uncertainty and disagreement. Let $\bar{x}_{i}^k =  \frac{\sum_{t=1}^{T_i^k} x_i^k(t)}{T_i^k}$ represent the average judgement from all OOB models. The right hand side of the equation can be further broken down as follows:

\begin{align}
\label{eq:num_two_terms}
S_i^k & = \frac{1}{T_i^k} \sum_{t=1}^{T_i^k} (\hat{x}_i^k(t) - x_i^k  )^2 \nonumber  \\
&= \frac{1}{T_i^k} \sum_{t=1}^{T_i^k} (\hat{x}_i^k(t) -   \bar{x}_{i}^k  + \bar{x}_{i}^k    - x_i^k  )^2 \nonumber \\
& = \frac{1}{T_i^k} \sum_{t=1}^{T_i^k} (\hat{x}_i^k(t) -   \bar{x}_{i}^k ) ^2 \nonumber  \\  - 
&\frac{2}{T_i^k} \sum_{t=1}^{T_i^k}   (\hat{x}_i^k(t) -   \bar{x}_{i}^k ) (\bar{x}_{i}^k    - x_i^k)  + (\bar{x}_{i}^k    - x_i^k ) ^2.
\end{align}
The first term, $\frac{1}{T_i^k} \sum_{t=1}^{T_i^k} (\hat{x}_i^k(t) -   \bar{x}_{i}^k ) ^2$,  is the variance of the predictions of the OOB models, which provides a measure of uncertainty. We can easily show that the middle term is equal to 0 since $\sum_{t=1}^{T_i^k}  \hat{x}_i^k(t) $ and  $\sum_{t=1}^{T_i^k}\bar{x}_{i}^k $ are equal. The last term represents the disagreement between the average OOB judgement and the actual feature value. Therefore, for a numerical feature, the anomaly can also be measured in aspects of uncertainty and disagreement as follows.  

\begin{equation}
S_i^k(\textrm{uncertainty}) =  \frac{1}{T_i^k} \sum_{t=1}^{T_i^k} (\hat{x}_i^k(t) -   \bar{x}_{i}^k ) ^2
\end{equation}
\begin{equation}
S_i^k(\textrm{disagreement}) =   (\bar{x}_{i}^k    - x_i^k ) ^2 = \left( \frac{\sum_{t=1}^{T_i^k} x_i^k(t)}{T_i^k} - x_i^k\right)^2.
\end{equation}

\subsection{Score Standardization}
Each dataset may consist of both categorical and numerical features, and each feature and its OOB predictions could follow different distributions. As a result, the anomaly scores computed from different features may have different scales. In order to combine the anomaly scores from different features, disregarding their types and distributions, we applied a min-max scaling to each anomaly score before combining them into a final score in \cref{eq:agg_func_avg}. 

\begin{equation}
S^k_{min}  = min(S^k_1, \ldots, S^k_N), k = 1, \ldots, K \\
\end{equation}
\begin{equation}
S^k_{max} = max(S^k_1, \ldots, S^k_N), k = 1, \ldots, K
\end{equation}
\begin{equation}\label{eq:min_max}
S^{k * }_i  = \frac{S^k_i - S^k_{min}}{S^k_{max} - S^k_{min}}.
\end{equation}

\begin{algorithm}
 \caption{OOB Anomaly Detection}
 \label{alg:algorithm}
\begin{algorithmic}[1]
  \STATE {\bfseries Input:} Dataset $\{x_i^k \}_{i = 1, \ldots, N, k= 1, \ldots, K}$ consisting of $N$ data points each with $K$ features 
  \STATE {\bfseries Output:} anomaly scores $S_i$ for $i=1,2, \ldots, N$
  \FOR{$k=1$ \bfseries{to} $K$}
    \STATE Train an ensemble model  $\bold{M}^k: \bold{x}^{-k} \rightarrow \bold{x}^k$
    \STATE With model $\bold{M}^k$, generate OOB predictions $ \hat{\bold{x}}_i^k = \{ \hat{x} _i^k(1),  \ldots,  \hat{x} _i^k(T_i^k) \} $ for $x_i^k, i = 1, \ldots, N$
    \STATE Compute $S_i^k , i=1, \ldots, N$ (\cref{eq:two_terms})
    \STATE Perform min-max scaling on $S_i^k$  (\cref{eq:min_max})
  \ENDFOR
\FOR{$i=1$ \bfseries{to} $N$}
  \STATE $S_i = \sum_{k=1}^{K} S_i^k$
\ENDFOR
\end{algorithmic}
\end{algorithm}

\subsection{Reasons for uncertainty and disagreement}
Here we provide insights into why we consider uncertainty and disagreement two good measures for detecting anomalies. We begin the discussion with two common types of anomaly:

The first type of anomaly is \textbf{outlier}, namely the data points that fall outside the the usual patterns of the dataset. Consequently, the OOB models have very little knowledge learnt from the remainder of the dataset on how to correctly predict them and therefore make contradicting predictions. This phenomenon often can be captured by a high uncertainty.

The second type is \textbf{mislabelled data}. This situation occurs when only one or very few features out of many were mistakenly recorded but their true value could be easily inferred from the remaining features. 
In real estate transactions, one example is human error in entering numbers to computer systems that may result in an extra or a missing '0' in the recorded sale price. Another example is a "townhouse" listing being mistakenly recorded as a "single family house". In those cases, the OOB predictions could have low uncertainty but high disagreement with the observed feature value, where the OOB predictions concentrate on a different feature value.

\begin{figure}
\begin{center}
  \centerline{\includegraphics[width=0.8\textwidth]{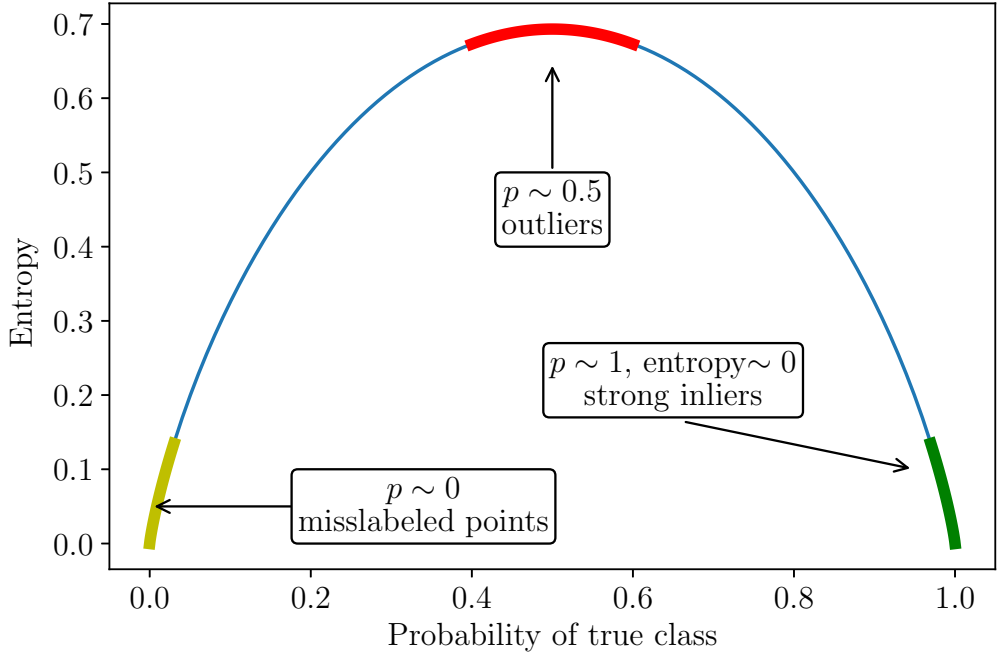}}
   \caption{
          Possible entropy values in light blue vs. OOB  probability of the observed feature value in the case of a 2-cardinality categorical. The red region characterizes outliers while the gold region characterizes the mislabelled data points.}
  \label{fig:entropy_sketch_2}
\end{center}
\end{figure}

\begin{figure}
\begin{center}
  \centerline{\includegraphics[width=0.8\textwidth]{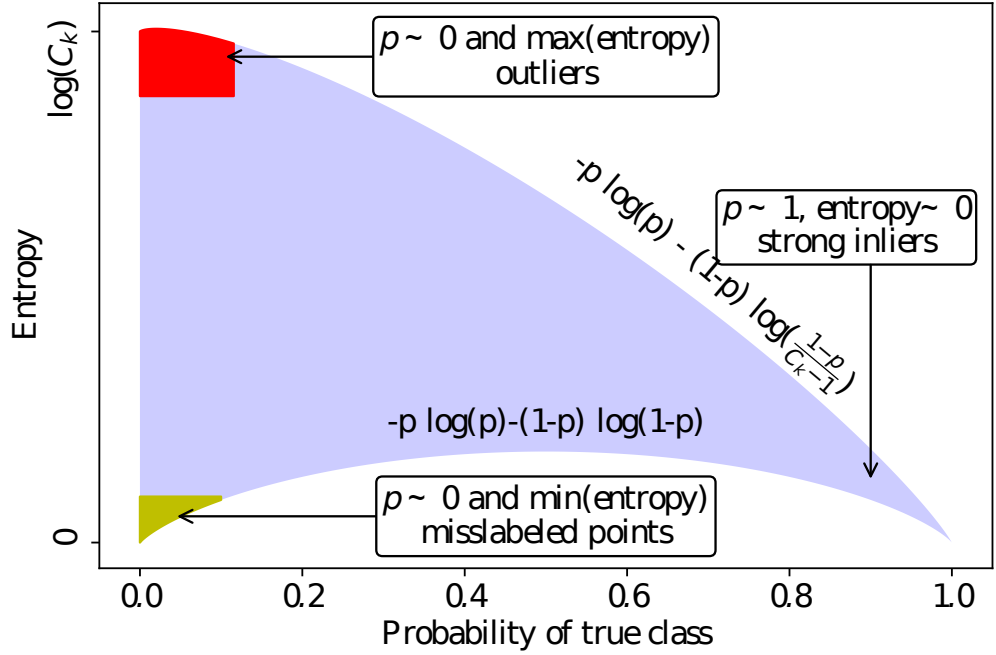}}
   \caption{
          Region of possible entropy values in light blue vs. OOB probability of the observed feature value in the case of $C_k$-cardinality categorical feature. The red region characterizes outliers, and the gold region characterizes the mislabelled data points.}
  \label{fig:entropy_sketch_n}
\end{center}
\end{figure}

We take categorical features as examples for a further visual illustration, but a similar intuition should generalize to numerical features. Fig.~\ref{fig:entropy_sketch_2} depicts the simple case when $C_k$=2, and~\cref{fig:entropy_sketch_n} depicts a more general scenario when $C_k$>2. In both figures, the x-axis represents the probability mass on the observed feature value from the OOB predictions, which we denote as  $p =\frac{\sum_{t=1}^{T_i^k} \bold{I}(x_i^k(t) == x_i^k)}{T_i^k}$ for simplicity, and the y-axis represents the entropy-based uncertainty measure. Fig.~\ref{fig:entropy_sketch_n} also sketches out the upper and lower bounds of the uncertainty measure. For a given value of $p$, the lower bound curve on the entropy occurs when one of the remaining $C_k-1$ classes carries all of the remaining mass $1-p$ (\cref{eq:lower_bound}), while the upper bound curve is reached when the remaining mass $1-p$ evenly distributed among the remaining $C_k$ classes 
(\cref{eq:upper_bound}). 
In both figures, the red regions are the outlier points which are characterized with high uncertainty in the OOB predictions. The gold regions are occupied by the possible mislabelled data points, where the OOB predictions are concentrated on a different categorical feature value with very little uncertainty. The lower right corner of the figures depicts the characteristics of the strong inliers, where the observed feature value is unsurprising and the OOB models are able to predict them correctly with high confidence: 
\begin{equation}\label{eq:lower_bound}
S_{lower}(\textrm{uncertainty}) =-p\ \text{log}(p) - (1-p)\   \text{log}(1-p),
\end{equation}
\begin{equation}\label{eq:upper_bound}
S_{upper}(\textrm{uncertainty}) =-p\ \text{log}(p) - (1-p)\ \text{log}\left(\frac{1-p}{C_k-1}\right).
\end{equation}

\subsection{Computational complexity}
\label{subsec:o_n}
The main computational cost is the training and scoring of each feature with the ensemble model. Training a random forest model with $T$ trees on a dataset with $K$ feature and $N$ data points is of $O(\sqrt{K} TN \log N)$, and  scoring the same size of data will cost $O(TN \log N)$. The algorithm requires such training and scoring process for each of the K features. Therefore, the overall computational complexity is $O(K^{3/2} TN \log N)$, which converges to $O(N\log N)$ when $N \gg K, T$. For real-life applications where important features can be identified by domain knowledge, we could potentially reduce the computational time by focusing the subset of important features instead of all features.

\subsection{Hyper-parameters}
\label{subsec:hyper_parameters}
Our method introduces a few standard parameters during the training of random forest ensemble models. 500 trees are trained for each Random Forest model. As the stopping criterion, the minimum leaf node size is set to be around 4\% of the size of datasets. We apply the categorical feature scoring method in \cref{sec:cat_feature} to the features that have significant smaller unique values than the size of the datasets (< 5\%), and apply the numerical feature scoring method in \cref{sec:num_feature} to the remaining features.

\section{Experiments}
\label{sec:results}
We conducted two set of experiments to demonstrate the effectiveness of our method in anomaly detection across various domains. In the first set of experiments, we evaluated the performance of anomaly detection methods through 10 publicly available datasets from the Outlier Detection Data Sets (ODDS) library \citep{Rayana2016}. All the datasets were originally available in UCI machine learning repository \citep{asuncion2007uci}, but converted for anomaly detection experiments. The anomalous data points are either labelled or defined by the minority label class. In the second experiment, we used our in-house home valuation dataset as a case study to illustrate the benefits of incorporating anomaly detection into a real-world ML system.

\subsection{Anomaly Detection on Benchmark Datasets}

\begin{table*}[h]
 \caption{Comparison of 10 anomaly detection methods over 10 benchmark datasets on AUC, average rank of AUC and Wilcoxon signed-rank test}
    \label{tab:uci_auc}
\resizebox{\textwidth}{!}{%
\begin{tabular}{|c|c|c|c|c|c|c|c|c|c|c|c|c|c|}
\hline
Data          & \# Samples    & \# Dim.    & Outlier \%   & KNN(5)          & KNN(10) & ABOD   & HBOS            & PCA    & VAE    & AE              & OCSVM  & IF              & OOB             \\ \hline
arrhythmia   & 452           & 274        & 14.60        & 0.7684          & 0.7746  & 0.7364 & 0.8141          & 0.7748 & 0.7748 & 0.7767          & 0.7737 & 0.8041          & \textbf{0.8196} \\ \hline
glass         & 214           & 9          & 4.21         & \textbf{0.8325} & 0.8054  & 0.7182 & 0.7057          & 0.6027 & 0.5986 & 0.5718          & 0.5496 & 0.7027          & 0.7927          \\ \hline
ionosphere    & 351           & 33         & 35.90        & 0.9337          & 0.9292  & 0.9170 & 0.5661          & 0.7947 & 0.7955 & 0.8184          & 0.8510 & 0.8556          & \textbf{0.9455} \\ \hline
optdigits     & 5216          & 64         & 2.88         & 0.3948          & 0.3724  & 0.5133 & 0.8723          & 0.5137 & 0.5137 & 0.5128          & 0.5071 & 0.7066          & \textbf{0.9484} \\ \hline
pima          & 768           & 8          & 34.90        & 0.7087          & 0.7134  & 0.6669 & 0.7089          & 0.6322 & 0.6587 & 0.6096          & 0.6237 & 0.6677          & \textbf{0.7161} \\ \hline
satellite     & 6435          & 36         & 31.64        & 0.6650          & 0.6872  & 0.5489 & \textbf{0.7557} & 0.6012 & 0.6012 & 0.6017          & 0.6636 & 0.7055          & 0.7462          \\ \hline
satimage-2    & 5803          & 36         & 1.22         & 0.9318          & 0.9584  & 0.7588 & 0.9767          & 0.9772 & 0.9772 & 0.9772          & 0.9967 & 0.9930          & \textbf{0.9981} \\ \hline
shuttle       & 49097         & 9          & 7.15         & 0.6342          & 0.6582  & 0.6142 & 0.9843          & 0.9899 & 0.9898 & 0.9898          & 0.9917 & \textbf{0.9968} & 0.9816          \\ \hline
vertebral     & 240           & 6          & 12.50        & 0.3781          & 0.3537  & 0.3652 & 0.3051          & 0.3776 & 0.3819 & \textbf{0.4684} & 0.4197 & 0.3542          & 0.3977          \\ \hline
vowels        & 1456          & 12         & 3.43         & \textbf{0.9768} & 0.9694  & 0.9661 & 0.6766          & 0.6062 & 0.6217 & 0.6009          & 0.7784 & 0.7547          & 0.9211          \\ \hline
\multicolumn{4}{|c|}{Avg. rank}                           & 5.3             & 5.2     & 7.0    & 5.1             & 6.4    & 6.7    & 6.7             & 5.8    & 4.4             & 2.4             \\ \hline
\multicolumn{4}{|c|}{Wilcoxon signed-rank test $p$ value} & 0.0372          & 0.0297  & 0.0063 & 0.0109          & 0.0035 & 0.0035 & 0.0109          & 0.0109 & 0.0047          & -               \\ \hline
\end{tabular}%
}
\end{table*}

The 10 ODDS benchmark datasets are good representations of real-world problems from various domains. Their sizes vary from hundreds of data points to hundreds of thousands of data points, and their number of features ranges from 6 to over 270. Given the known labels for anomalies, we used the Area Under the Curve (AUC) of Receiver Operating Characteristics (ROC) to evaluate the performance of the anomaly detection methods. ROC was calculated by sweeping over different thresholds on the anomaly scores, and counting the number of true positives and false negatives, given that anomalies are labelled as positive. 

In this set of experiments, we consider the following competitors, which cover existing methods in all 4 categories as summarized in \cref{sec:related}:
\begin{itemize}
    \item KNN: K Nearest Neighbors (use the distance to the kth nearest neighbor as the outlier score, with K=5 and 10) \citep{ramaswamy2000efficient}
    \item ABOD: Angle-Based Outlier Detection \citep{kriegel2008angle}
    \item HBOS: Histogram-Based Outlier Score \citep{goldstein2012histogram}
    \item PCA: PCA-based anomaly detection \citep{shyu2006principal}
    \item AE: AutoEncoder (use reconstruction error as the outlier score) \citep{kramer1991nonlinear}
    \item VAE: Variational AutoEncoder (use reconstruction error as the outlier score) \citep{kingma2013auto}
    \item OCSVM: One-Class Support Vector Machine \citep{scholkopf2000support}
    \item IF: Isolation Forest \citep{isolation_forest}
\end{itemize}

In our experiments, we used the implementations with default paramters in \citep{zhao2019pyod} for these competitors. For AE and VAE, the encoder hidden layers have 16, 8 and 4 nodes for datasets with more than 16 features and 8, 8 and 4 nodes for the remaining datasets, and the decoders follow a reversed setting. 

For each method and each dataset, the average AUCs over 10 experiments are reported in \cref{tab:uci_auc} to account for the stochastic processes in the methods. For each dataset, the highest average AUC achieved is highlighted in bold, and we rank the methods from 1 to 10 based on their AUC performance. The average rank for each method across 10 datasets is also reported in \cref{tab:uci_auc}.

From \cref{tab:uci_auc}, we observe that the proposed OOB anomaly detection method outperforms all 9 competing methods in 5 out of 10 datasets. Moreover, our method achieves the highest average rank (2.4), while the second best method, IF \citep{isolation_forest}, achieves the average rank of 4.4. In addition, we use Wilcoxon signed-rank test \citep{wilcoxon1992individual} to compare each competing method to our method with significance measurement. AUC metrics from two methods for the same dataset are considered paired observations when computing the test statistics, and the alternative hypothesis is that the AUC performance of OOB method is greater than that of the competing method. Based on the results of the Wilcoxon signed-rank test reported in the final row of \cref{tab:uci_auc}, our proposed method is better than each of the competing methods at significance level 0.05. This set of experiments demonstrates the strong performance of our method across a diverse set of anomaly detection tasks.

\subsection{Case Study: Anomaly Detection for Automatic Home Valuation}
\label{sec:zillow_data}

Motivated by the home valuation application at \textsc{Zillow}, we designed a case study to illustrate how a ML system could benefit from the OOB anomaly detection method. We used an in-house dataset originally constructed to perform a home valuation task as a regression problem. The dataset contains near $2.1 * 10^6$ data points (historical real estate transactions) from a city region with a mixture of categorical, ordinal, and continuous data types. In this experiment, we compare our method with only IF, the strongest competitor based on the experiments on ODDS datasets.

\begin{figure}[t]
\begin{center}
    \includegraphics[width=0.8\textwidth]{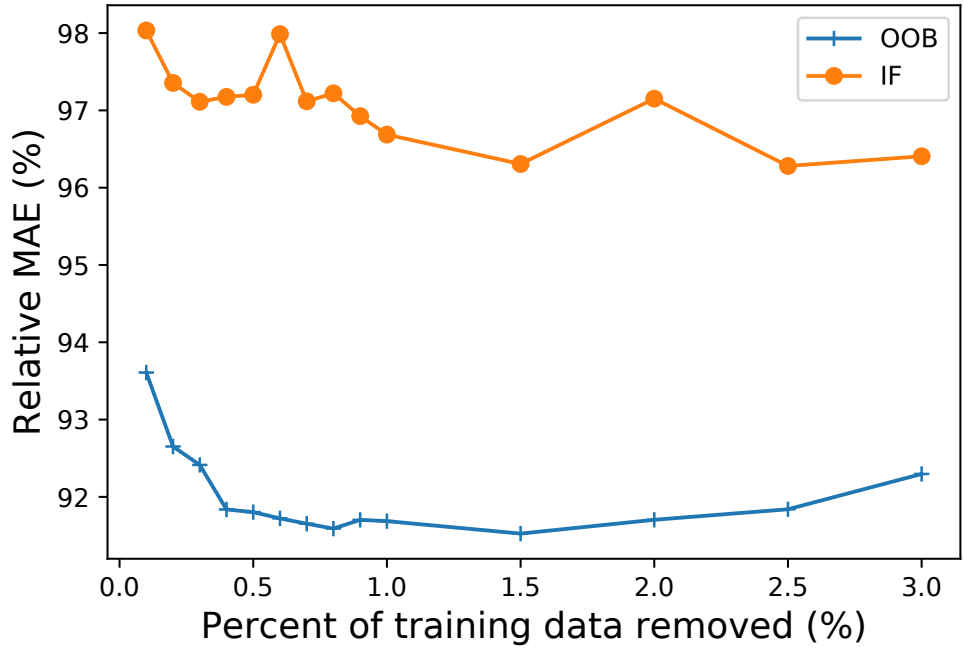}
    \caption{Comparison of the proposed method and Isolation Forest (IF) on the case study of home valuation modeling. For each method, a curve shows the relative MAEs as we remove different percentages of training data labelled as anomalies}
    \label{fig:zillow_error}
\end{center}
\end{figure}

Although we don't have ground truth for the anomalies, our goal of integrating anomaly detection as a pre-processing step is to improve the accuracy of the ML models, which determines the following evaluation approach. We first split the data into a training dataset (80\%) and a testing dataset (20\%). We evaluated different anomaly detection methods based on the prediction accuracy of the testing data after cleaning the training data with these methods. More specifically, we first remove a certain percentage of data with the highest anomaly scores provided by an anomaly detection model, then train a home valuation model (based on LightGBM \citep{ke2017lightgbm}) with the cleaned training data after removing these anomalies, and finally evaluate the model performance on the hold-out testing set. The relative mean absolute error (MAE) after removing different percentages of training data points are reported in \cref{fig:zillow_error}. The relative MAE is calculated as the ratio to the original MAE achieved by training on the full training dataset without any anomaly removal.

From \cref{fig:zillow_error}, we observe that both IF and OOB are able to reduce the testing error by excluding a small percentage of training data from the training process. This observation highlights the importance of anomaly detection in an ML system. For Zillow, any small error reduction in home valuation can have a great positive impact in business objectives and obtaining customer trust.

Furthermore, we observe a consistent advantage of OOB anomaly detection method over IF in this case study, which strongly demonstrates the effectiveness of our method. We also observe a slight bounce of error after removing 1.5\% of training data for both methods, which may be an indicator of removing too much (valid) training data. Therefore, before integrating any anomaly detection methods into a production data processing pipeline, an offline evaluation similar to
\cref{fig:zillow_error} should be conducted to determine a proper anomaly score threshold (based on the ``elbow'' points).

\section{Conclusions and Future Work}
\label{sec:conclusion}

In this paper, we introduced a novel method for unsupervised anomaly detection, which we called Out-of-Bag anomaly detection. Our method decomposes the unsupervised problem into a set of supervised learning problems and leverages the statistics derived from the OOB estimates. The proposed method is applicable to complex datasets with mixed categorical and numerical features. We demonstrated effectiveness of our anomaly detection method through experiments on standard benchmark datasets and an in-house case study on home valuation.

For future research, a possible improvement to our method is combining feature selection or weighting into the current algorithm. An algorithmic way to identify the most important feature for anomaly detection may further improve the effectiveness and efficiency of our method. For example, for a housing dataset, the anomaly score derived from the price feature may be more important than scores from other features. Also, it will be a valuable investigation to explore classification or regression models other than decision trees and regression trees. Although we used tree-based models in this paper, the proposed OOB framework is general enough and could be extended to use other base learners. 

For future applications, in addition to using our method in data pre-processing as discussed in \cref{sec:zillow_data}, our method can also be used to detect fraudulent listings, which is a very important topic for housing marketplaces like Zillow, Trulia and Redfin as well as for travel websites like Airbnb and Vrbo. In the future, we will also explore how this unsupervised anomaly signal can be leveraged in our ranking and recommendation systems in order to better protect our users.

Anomaly detection methods can help ML applications in achieving reliable and robust performance in the domains where customers trust is highly valued. Here at Zillow, adopting anomaly detection early on in ML applications has helped us improve the accuracy of ML models and build a trusted home-related marketplace \citep{zillow_end_game}. We hope our method can also provide values for other customer-centric machine leanring groups.

We have made a Python implementation
of the OOB anomaly detection algorithm publicly available\footnote{{https://github.com/comorado/OOB\_anomaly\_detection}}, accompanied by an example using the Jupyter notebooks for reproducibility of the results reported in this work. For training OOB models we use \textsc{scikit-learn} library \cite{scikit-learn}.

\nocite{*}
\bibliographystyle{ACM-Reference-Format}
\bibliography{oob_outlier_detection}
\end{document}